%% file: aaai24.tex
\documentclass[letterpaper]{article} 
\usepackage{aaai24}  
\usepackage{times}  
\usepackage{helvet}  
\usepackage{courier}  
\usepackage[hyphens]{url}  
\usepackage{graphicx} 
\usepackage{natbib}  
\usepackage{caption} 
\usepackage{algorithm}
\usepackage{comment}
\usepackage[noend]{algpseudocode}
\newtheorem{Theorem}{Theorem}
\newtheorem{Definition}{Definition}
\usepackage{dsfont}
\usepackage{xcolor}

\usepackage{booktabs}
\usepackage{multirow}
\usepackage{multicol}
\usepackage{subfigure}
\usepackage{amsmath}
\usepackage{amssymb}
\usepackage{mathtools}
\usepackage{tikz}
\usepackage{newfloat}
\usepackage{listings}
\DeclareCaptionStyle{ruled}{labelfont=normalfont,labelsep=colon,strut=off} 
\floatstyle{ruled}
\newfloat{listing}{tb}{lst}{}
\floatname{listing}{Listing}
\title{ReRoGCRL: Representation-based Robustness in Goal-Conditioned Reinforcement Learning}
\author {
    Xiangyu Yin\textsuperscript{\rm 1}\equalcontrib,
    Sihao Wu\textsuperscript{\rm 1}\equalcontrib,
    Jiaxu Liu\textsuperscript{\rm 1},
    Meng Fang\textsuperscript{\rm 1},
    Xingyu Zhao\textsuperscript{\rm 2},\\
    Xiaowei Huang\textsuperscript{\rm 1},
    Wenjie Ruan\textsuperscript{\rm 1}\footnote{Corresponding Author}
}
\affiliations {
    \textsuperscript{\rm 1} Department of Computer Science, University of Liverpool, Liverpool, L69 3BX, UK\\
    \textsuperscript{\rm 2} WMG, University of Warwick, Coventry, CV4 7AL, UK\\
    \{X.Yin22;sihao.wu;Meng.Fang;xiaowei.huang\}@liverpool.ac.uk, xingyu.zhao@warwick.ac.uk, w.ruan@trustai.uk
}

\usepackage{bibentry}

\begin{document}

\maketitle

\begin{abstract}
\input{Contents/abstract}
\end{abstract}

\input{Contents/Introduction}

\input{Contents/relatedworks}

\input{Contents/preliminaries}

\input{Contents/method}

\input{Contents/experiments}

\input{Contents/conclusion}

\input{Contents/acknowledgement}

\bibliography{aaai24}

\input{Contents/appendix}
\end{document}

%% file: Contents/abstract.tex
While Goal-Conditioned Reinforcement Learning (GCRL) has gained attention, its algorithmic robustness against adversarial perturbations remains unexplored. The attacks and robust
representation training methods that are designed for traditional RL become less effective when applied to GCRL. 
To address this challenge, we first propose the \textit{Semi-Contrastive Representation} attack, a novel approach inspired by the adversarial contrastive attack. Unlike existing attacks in RL, it only necessitates information from the policy function and can be seamlessly implemented during deployment. Then, to mitigate the vulnerability of existing GCRL algorithms, we introduce \textit{Adversarial Representation Tactics}, which combines \textit{Semi-Contrastive Adversarial Augmentation} with \textit{Sensitivity-Aware Regularizer} to improve the adversarial robustness of the underlying RL agent against various types of perturbations. Extensive experiments validate the superior performance of our attack and defence methods across multiple state-of-the-art GCRL algorithms. Our tool {\bf ReRoGCRL} is available at \url{https://github.com/TrustAI/ReRoGCRL}.

%% file: Contents/introduction.tex
\section{Introduction}

Goal-Conditioned Reinforcement Learning (GCRL) trains an agent to learn skills in the form of reaching distinct goals. Unlike conventional RL, GCRL necessitates the agent to make decisions that are aligned with goals. This attribute allows agents to learn and accomplish a variety of tasks with shared knowledge, better generalization, and improved exploration capabilities. Furthermore, it is binary-rewarded, which is easier to implement compared to hand-crafted complex reward functions. Recently, there has been a significant surge in research related to GCRL. Exemplary works include methods based on techniques such as hindsight experience replay \cite{andrychowicz2018hindsight, fang2018dher, fang2019curriculum}, imitation learning \cite{ghosh2020learning, yang2021rethinking}, or offline learning \cite{chebotar2021actionable, ma2022offline, mezghani2023learning}.

Generally, these studies tend to assume that the sensing and perception systems of agents are devoid of uncertainties. However, this presumption is hardly applicable in real-world scenarios because there are known gaps between simulated and real-world environments, such as measurement errors, motor noise, etc. The observations made by agents encompass unavoidable disturbances that originate from unforeseeable stochastic noises or errors in sensing \cite{huang2020survey,zhang2023reachability,wang2022deep,huang2012deep}. For example, due to the existence of adversarial attacks in RL agents \cite{huang2017adversarial, weng2019toward, bai2019model,mu2023reward,mu2023certified}, even minor perturbations can lead to unsafe outcomes in safety-critical GCRL control strategies~\cite{gleave2019adversarial,sun2020stealthy}. 

Recently, numerous robust techniques have been implemented in traditional RL, which can be broadly classified into two categories: {\em i)} Adversarial Training (AT), and {\em ii)} Robust Representation Training. Particularly, AT in RL trains agents with adversarial states or actions \cite{pinto2017robust, kos2017delving, zhang2020robust}, and then enhances their adversarial robustness. However, AT-trained RL agents cannot be utilized across downstream tasks, which limits their transferability. Compared to AT, robust representation training can deliver \textit{low-dimentional}, \textit{collapse-resistant}, and \textit{perturbation-robust} representations of observations \cite{gelada2019deepmdp, zhang2020learning, zang2022simsr}, which is the focus of recent studies. Specifically, it learns representations that capture only task-relevant information based on the bisimulation metric of states \cite{ferns2011bisimulation}. Technically, this is achieved by reducing the \textit{behavioural difference} between representations of similar observation pairs in the latent space.

Although there are already many articles enhancing the robustness of RL, few works \cite{10018434} consider the robustness of GCRL against adversarial perturbations.
Different from the vanilla one, GCRL employs reward functions characterized by sequences of unshaped binary signals. For example, \cite{andrychowicz2018hindsight} allocates the reward by determining whether the distance between the achieved goal $\hat{g}$ and the desired goal $g$ is less than a threshold $\epsilon$: $\mathds{1}(\left\|\hat{g}-g\right\|\leq\epsilon)$. This makes the reward sequences in GCRL notably sparser than those in traditional RL. 
The sparsity of rewards in GCRL leads to inaccurate estimations of both Q-values and actions. This presents a significant challenge for directly implementing conventional RL attacks, particularly those that rely on pseudo labels such as Q-values or action values.
Consequently, this underscores the necessity to develop new attack methods tailored for GCRL to evaluate its robustness. Inspired by adversarial contrastive learning \cite{kim2020adversarial, jiang2020robust, ho2020contrastive}, we propose the Semi-Contrastive Representation (SCR) attack, with the aim of maximizing the distance between the representations of original states and their corresponding perturbed versions. This approach effectively bypasses the need for approximated labels. 
Notably, this method can operate independently of the critic function and is readily deployable.

\begin{figure}[tb]
\centering
\includegraphics[width=0.45\linewidth]{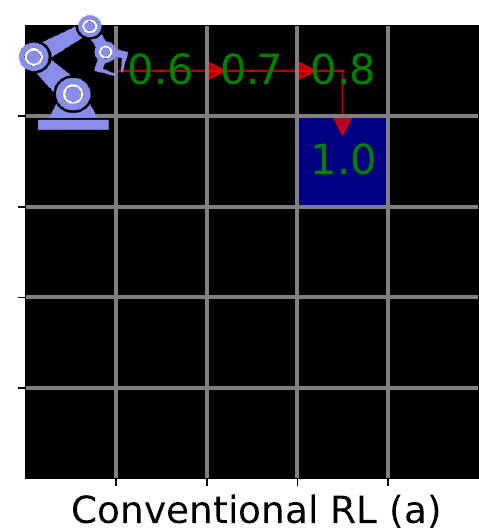}
\includegraphics[width=0.45\linewidth]{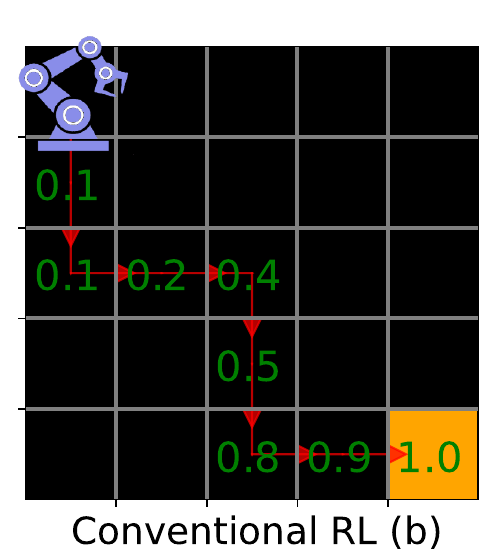}

\vspace{1mm}

\includegraphics[width=0.45\linewidth]{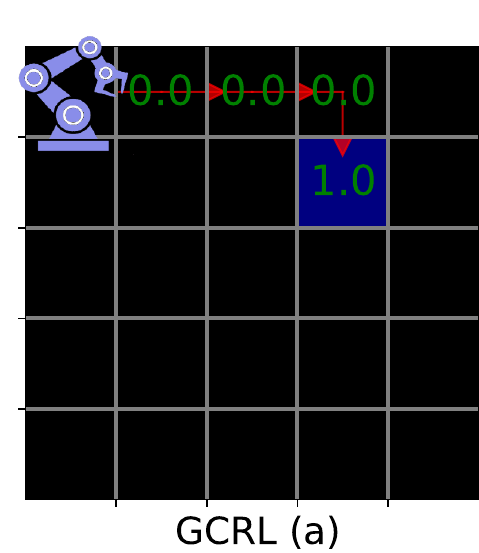}
\includegraphics[width=0.45\linewidth]{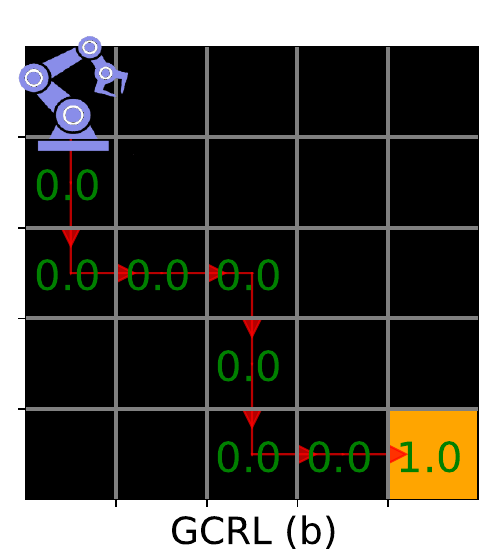}
\caption{Trajectories of the agent at state $s$ approaching \textcolor{blue}{blue} and \textcolor{orange}{orange} goals in conventional RL and GCRL, where the designated goals vary with different initialization. Rewards are indicated in each block along the trajectories.}
\label{figure:special_case}
\end{figure}

Contrary to conventional RL, where state rewards approach 1.0 as the agent nears the goal, in GCRL, rewards consistently remain at 0.0 until the goal is reached. This distinction is highlighted in Fig.~\ref{figure:special_case}: for a given state $s$, the rewards of subsequent states in Conventional RL (a) significantly differ from those in Conventional RL (b). However, such a disparity in rewards is not present in GCRL, where they invariably stay at 0.0. Consequently, bisimulation metric-based representation training methods cannot capture well the differences between state-goal tuples in GCRL. To deal with this, we propose a universal defensive framework named Adversarial Representation Tactics (ARTs). Firstly, ARTs enhance the robustness of vanilla GCRL algorithms with Semi-Contrastive Adversarial Augmentation (SCAA). Secondly, ARTs mitigate the performance degradation associated with bisimulation metric-based robust representation techniques using the Sensitivity-Aware Regularizer (SAR).
In summary, our contributions are summarized as follows:
\begin{itemize}

\item We introduce a novel Semi-Contrastive Representation (SCR) attack. It can operate without the critic function and is ready for direct deployment.

\item We propose a mixed defensive strategy, termed Adversarial Representation Tactics (ARTs), to dynamically enhance adversarial robustness tailored to specific GCRL algorithms.

\item Extensive experiments validate that our proposed attack method and defence techniques outperform state-of-the-art algorithms in GCRL by a large margin.
\end{itemize}

%% file: Contents/relatedworks.tex
\section{Related Work}
\subsection{Goal-Conditioned Reinforcement Learning}

Several methods have tackled GCRL. Hindsight experience replay \cite{andrychowicz2018hindsight} is
often employed to enhance the effectiveness of policies by relabeling the goal. Goal-conditioned supervised learning approaches \cite{ghosh2020learning, yang2021rethinking} solve this problem by iteratively relabeling and imitating self-generated experiences. For long-horizon tasks, hierarchical RL \cite{chane2021goal} learns high-level policies to recursively estimate a sequence of intermediate subgoals. Moreover, model-based methods \cite{charlesworth2020plangan, yang2021mher}, self-supervised learning \cite{mezghani2022learning, eysenbach2022contrastive}, and planning-based methods \cite{eysenbach2019search, nasiriany2019planning} are also used to solve GCRL problems.
Unlike these prior methods, our paper considers the representation training and robustness performance on GCRL. 

\subsection{Adversarial Contrastive Attack}
Adversarial contrastive attacks have gained significant attention in recent studies. Particularly, \citep{kim2020adversarial} proposes the instance-wise adversarial attack for unlabeled data, which makes the model confuse the instance-level identities of the perturbed data samples. \citep{jiang2020robust} directly exploits the Normalized Temperature-scaled Cross Entropy loss to generate PGD attacks. \citep{ho2020contrastive} introduces a new family of adversarial examples by backpropagating the gradients of the contrastive loss to the network input. While the above methodologies employ varied methods to generate adversaries, they consistently rely on the cosine similarity among the original sample, positive, and negative samples. In this paper, however, we devise a novel contrastive attack method, which is more suitable in GCRL and can be extended to other RL settings.

\subsection{Bisimulation Metric}
Bisimulation metrics offer a framework to gauge the similarity between states within MDPs. The traditional bisimulation metric, as defined by \cite{ferns2012metrics}, considers two states to be close if their immediate rewards and transition dynamics are similar \cite{larsen1989bisimulation, givan2003equivalence}. This self-referential concept has been mathematically formalized using the Kantorovich distance, leading to a unique fixed-point definition. However, the conventional approach has been criticized for resulting in \textit{pessimistic} outcomes, as it requires consideration of all actions, including those that may be suboptimal. To address this challenge, \cite{castro2022mico} introduced the strategy of Matching Under Independent Couplings (MICo). This approach focuses solely on actions induced by a specific policy $\pi$, offering a more optimistic and tailored measure of state equivalence. Whereas, in this work, we focus on their extended version, Simple State Representation (SimSR).

%% file: Contents/preliminaries.tex
\section{Background}
\label{preliminaries}
In this section, we first present the fundamental concepts of goal-conditioned MDPs and delve into their extensions in adversarial state-goal scenarios. Then, we outline the architecture of our training backbone and describe SimSR, the cutting-edge robust representation training method.
\subsection{Goal-Conditioned MDPs}
In this section, we model our task as a goal-conditioned MDP. This can be formalized as a tuple $(\mathcal{S}, \mathcal{G}, \mathcal{Z}, \mathcal{A}, r, \gamma, \mathcal{T})$. Here, $\mathcal{S}$ represents the state space, $\mathcal{G}$ the goal space, $\mathcal{Z}$ the latent representation space, and $\mathcal{A}$ the action space. The reward function is denoted by $r$, $\gamma$ stands for the discount factor, and $\mathcal{T}$ signifies the state-goal transition probability function.

We assume that the policy function $\pi$ can be approximated using a combination of an encoder, $\psi(\cdot)$, and an actor-network, $\varphi(\cdot)$. The critic function is represented by $\varrho(\cdot)$. Specifically, for any given state $s \in \mathcal{S}$ and goal $g \in \mathcal{G}$, the encoder $\psi(\cdot)$ transforms the concatenated tuple $\left<s, g\right>$ into the representation space $\mathcal{Z}$. The resulting feature, $z_{\left<s,g\right>}$, is expressed as $\psi(\left<s, g\right>)$. The actor network, $\varphi(\cdot)$, maps this feature, $z_{\left<s,g\right>}$, to a specific action $a_{\left<s,g\right>}\in\mathcal{A}$. The action distribution is defined as $a_{\left<s,g\right>}\sim\pi(\cdot|\left<s, g\right>)$. Based on the chosen action $a_{\left<s, g\right>}$, the agent's subsequent state, $s^{\prime}$, can be sampled using $s^{\prime}\sim\mathcal{T}(\cdot|\left<s,g\right>, a_{\left<s,g\right>})$ and succinctly represented as $\mathcal{T}^{\pi}_{\left<s,g\right>}$.

In contrast to traditional RL algorithms, GCRL only provides rewards when the agent successfully achieves a predefined goal. Specifically, we define the reward for a state-goal tuple $\left<s, g\right>$ as $r_{\left<s,g\right>} = \mathds{1}\left(\mathcal{D}(s, g)\leq\eta\right)$. Here, $\eta$ is a predefined threshold, $\mathcal{D}$ acts as a distance metric, determining if $s$ and $g$ are sufficiently close. We utilize $\ell_\infty$-norm in this paper. Given an initial state $s_0$ and a goal $g$, our primary objective is to optimize the expected cumulative rewards across joint distributions following:
\begin{align}
\label{objective}
    J_{\pi}(s_0, g) = \mathbb{E}_{\substack{ a_t\sim\pi(\cdot|\left<s_t,g\right>),\\s_{t+1}\sim\mathcal{T}^{\pi}_{\left<s_t, g\right>}}}\left[\sum_{t=0}^{T-1}\gamma^{t}r_{\left<s_t, g\right>}\right].
\end{align}
In GCRL, the reward function yields only two possible outcomes: 0 or 1, which results in a notably sparser reward sequence $(r_{\left<s_0, g\right>}, \cdots, r_{\left<s_{T-1}, g\right>})$ over a $T$-step trajectory compared to other RL algorithms. While these binary rewards may not offer granular insights into the agent-environment interactions, they do guide the agent in a strategy where it initially distances itself from the goal before converging towards it. To derive the optimal policy $\pi^{*}$, we refine Eq.~(\ref{objective}) using the Q-value at the initial state $s_0$, then $Q_{\pi}(\left<s_0, g\right>, a_0)$ can be estimated through the Bellman equations:
\begin{equation}
\label{qvalue}
r_{\left<s_0, g\right>}
+\mathbb{E}_{s_1\sim\mathcal{T}^{\pi}_{\left<s_0, g\right>}}\max_{a_1\sim\pi(\cdot|\left<s_1, g\right>)}Q_{\pi}(\left<s_1, g\right>, a_1).
\end{equation}

\begin{Definition}[\cite{zhang2020robust}]
\label{stationary}
The adversarial version of a state $s$ can be defined as $\mathcal{V}(s;\theta)$. It is considered stationary, deterministic, and Markovian if its behaviour is solely determined by $s$ and the policy network, which is parameterized by $\theta$. For simplicity, we denote $\mathcal{V}(s;\theta)$ as $\mathcal{V}(s)$.
\end{Definition}

\subsection{State-Goal Adversarial Observations}
Building on Def.~\ref{stationary}, we consider states and goals separately and individually construct the set of adversarial states $\mathcal{B}_p^{\epsilon}(s)$ and goals $\mathcal{B}_{p}^{\epsilon}(g)$ using the State-Adversarial MDP (SA-MDP) framework:
\begin{align}
\begin{split}
&\mathcal{B}_{p}^{\epsilon}(s)\coloneqq\{\mathcal{V}(s) \mid \left\|\mathcal{V}(s)-s\right\|_p\leq\epsilon_s\},\\
&\mathcal{B}_{p}^{\epsilon}(g)\coloneqq\{\mathcal{V}(g) \mid \left\|\mathcal{V}(g)-g\right\|_p\leq\epsilon_g\},
\end{split}
\end{align}
where $\epsilon_s$ and $\epsilon_g$ are predefined thresholds of adversarial perturbations on states and goals respectively. $\mathcal{V}(\cdot)$ acts as a stationary and deterministic mapping, which transforms a clean input into its adversarial counterpart. Note that the adversary introduces perturbations solely to the state-goal observations. Consequently, even though the action is determined as $a\sim\pi(\cdot|\left<\mathcal{V}(s), \mathcal{V}(g)\right>)$,  transitions of the environment are still governed by the original state-goal pair $\left<s,g\right>$, instead of its perturbed counterpart $\left<\mathcal{V}(s),\mathcal{V}(g)\right>$. Due to uncertainties in the state-goal estimation, there is a potential for generating actions that are not optimal.

\subsection{Neural Network-based Backbone}
\label{basic_notation}
In this study, we utilize Multi-Layer Perceptrons (MLPs) of varying depths as the foundational architectures for both the policy function, represented as $\varphi(\psi(\cdot))$, and the critic function, $\varrho(\cdot)$. Specifically, for an encoder $\psi(\cdot)$ comprising $L$ layers, its output, $o_{L}$, can be articulated as $o_{L} = \mathbf{W}_L\phi\left(\mathbf{W}_{L-1}\cdots\phi(\mathbf{W}_1o_0)\right)$, with the stipulation that $L\geq2$. Here, $o_0$ refers to the input vector in its flattened form, $\phi(\cdot)$ represents the ReLU-based activation function, and $\mathbf{W}_{i}$ is the weight matrix associated with the $i$-th layer. To be simplified, we denote the parameters and the depths of $\psi(\cdot)$, $\varphi(\cdot)$, and $\varrho(\cdot)$ as $\theta_{\psi}$, $\theta_{\varphi}$, $\theta_{\varrho}$, and $L_{\psi}$, $L_{\varphi}$, $L_{\varrho}$, respectively. For clarity, we omit the bias vector present at each layer.

\subsection{Simple State Representation (SimSR)}
In this paper, SimSR functions as a metric within the representation space, enhancing our base methods in GCRL. Unlike the bisimulation metric \cite{ferns2012metrics} or MICo \cite{castro2022mico}, which employ the Wasserstein or diffuse metric to gauge the distance between two distributions, SimSR adopts a non-diffuse metric, offering a more relaxed approach. Take input tuples $\left<s_i, g_1\right>$ and $\left<s_j, g_2\right>$ from Fig.~\ref{figure:special_case} for instance, the measurement $\mathcal{M}(\cdot, \cdot)$ between their representations is constructed using the cosine distance (a $\ell_2$ normalized dot product distance) as the foundational metric, formulated as:
\begin{equation}
\resizebox{0.88\linewidth}{!}{
$\begin{aligned}
\mathcal{M}(\left<s_i, g_1\right>, \left<s_j, g_2\right>) = 1 - \frac{\psi(\left<s_i, g_1\right>)^\top\psi(\left<s_j, g_2\right>)}{\left\|\psi(\left<s_i, g_1\right>)\right\|_2\left\|\psi(\left<s_j, g_2\right>)\right\|_2}
\end{aligned}$}
\end{equation}
Note that SimSR shares the same fixed point as MICo. Its update operator can be defined in a manner analogous to MICo.
\begin{Theorem}[\cite{zang2022simsr}]
\label{simsr}
Given a policy function $\pi$, the SimSR operator $T^{\pi}$ which defines the state-goal tuple similarity between $\left<s_i, g_1\right>$ and $\left<s_j, g_2\right>$ can be updated as:
\begin{align}
\label{simsr_operator}
    &(T^{\pi}\mathcal{M})(\left<s_i, g_1\right>, \left<s_j, g_2\right>) =|r_{\left<s_i, g_1\right>} - r_{\left<s_j, g_2\right>}| + \\
    & \gamma\mathbb{E}_{s_{i+1}\sim \mathcal{T}_{\left<s_i, g_1\right>}^{\pi}, s_{j+1}\sim \mathcal{T}_{\left<s_j,g_2\right>}^{\pi}}\left[\mathcal{M}(\left<s_{i+1}, g_1\right>, \left<s_{j+1}, g_2\right>)\right],\nonumber
\end{align}
for all $\mathcal{M}:\left<\mathcal{S}\times\mathcal{G}\right>\times\left<\mathcal{S}\times\mathcal{G}\right>\rightarrow\mathbb{R}$.
\end{Theorem}
The second term in Eq.~(\ref{simsr_operator}) is derived through sample-based approximation. Using the SimSR operator, one can iteratively refine the representation by applying it to an arbitrarily initialized $\psi(\cdot)$.

Note that by using cosine distance as the foundational metric in SimSR, all derived state features are normalized to unit length. Furthermore, in contrast to the bisimulation metric which employs the Wasserstein distance, the SimSR operator achieves a computational complexity on par with the MICo operator.

%% file: Contents/method.tex
\section{Semi-Contrastive Representation Attack}
We present a representation-based adversarial attack for GCRL, which is independent of the critic function and can be seamlessly integrated in the deployment phase.
\subsection{Limitations of Traditional Attacks in GCRL}
Adversarial attacks on states lack the gradient information from labeled examples. To navigate these constraints, existing methods lean on various pseudo-labels, such as Q-values \cite{zhang2020robust, kos2017delving} or actions \cite{gleave2019adversarial, sun2020stealthy} to generate adversarial states. However, as illustrated in Eq.~(\ref{qvalue}), Q-value function is heavily influenced by the reward sequence. Moreover, many of these attacks necessitate access to value networks, making them unsuitable for direct deployment in the inference stage. To address these challenges, we introduce a novel attack method tailored for GCRL algorithms. This approach neither depends on specific pseudo-labels nor requires access to the critic network. Although crafted with GCRL's unique characteristics, it is versatile enough for broader applications of RL algorithms.

\subsection{Negative Tuple-Based Adversary}

To craft adversaries without relying on labels or pseudo-labels, several studies \cite{ho2020contrastive, jiang2020robust, kim2020adversarial} have introduced the concept of adversarial contrastive attack in the realm of unsupervised learning. Viewed through the lens of representation, an adversary $\mathcal{V}(x)$ for a clean input $x$ is designed to ensure that the feature $f(\mathcal{V}(x))$ diverges significantly from $f(x^{+})$ while converging towards the feature of a negative sample, $f(x^{-})$. Typically, these methods employ the NT-Xent loss to gauge the similarity among $f(\mathcal{V}(x))$, $f(x^{+})$, and $f(x^{-})$. This loss can be articulated as:
{\footnotesize
\begin{equation}
\resizebox{0.88\linewidth}{!}{
$\begin{aligned}
-\log\frac{\sum_{\{x^{+}\}}\exp\left(\mathbb{S}(\mathcal{V}(x), x^{+})\right)}{\sum_{\{x^{+}\}}\exp\left(\mathbb{S}(\mathcal{V}(x), x^{+})\right)+\sum_{\{x^{-}\}}\exp\left(\mathbb{S}(\mathcal{V}(x), x^{-})\right)},
\end{aligned}$}
\end{equation}}
where $\mathbb{S}(a,b)=\frac{f(a)^{\top}f(b)}{\tau}$, $\tau$ refers to the temperature hyperparameter. Normally, the positive set $\{x^{+}\}$ can be constructed using different types of data augmentation, such as rotation, color jittering, or scaling, while $\left\{x^{-}\right\}$ indicates the set of samples collected from other classes. 
In RL contexts, it is challenging to directly quantify positively correlated samples. Thus, in this section, we present an adversarial contrastive attack approach that exclusively focuses on the negative tuple, termed the Semi-Contrastive Representation (SCR) attack in Def.~\ref{scr}.
\begin{Definition}[SCR attack]
\label{scr}
Given a feature extraction function $f(\cdot)$ and the input tuple $\left<s,g\right>$, the semi-contrastive representation attack can be defined as follows:
\begin{equation}
\resizebox{0.88\linewidth}{!}{
$\begin{aligned}
\mathop{\arg\sup}_{\mathcal{V}(s)\in\mathcal{B}_{p}^{\epsilon}(s), \mathcal{V}(g)\in\mathcal{B}_{p}^{\epsilon}(g)}\mathbb{E}_{\left<s,g\right>^{-}}\left[\mathcal{L}_{lg}\left(-f(\left<\mathcal{V}(s), \mathcal{V}(g)\right>)^{\top}f(\left<s,g\right>^{-})\right)\right],
\end{aligned}$}
\end{equation}
where the logistic function $\mathcal{L}_{lg}(v)=\log(1+\exp(-v))$ for any $v\in\mathbb{R}$, $\left<s,g\right>^{-}$ indicates the negative tuple.
\end{Definition}
It is easy to derive that $\frac{\mathcal{L}_{lg}(v_1)+\mathcal{L}_{lg}(v_2)}{2}\geq\mathcal{L}_{lg}(\frac{v_1+v_2}{2})$, so we can apply the sub-additivity to the supremum of the values of $\mathcal{L}_{lg}$ by:
\begin{equation}
\resizebox{0.88\linewidth}{!}{
$\begin{aligned}
\label{subadd}
&\sup_{\mathcal{V}(s), \mathcal{V}(g)}\mathbb{E}_{\left<s,g\right>^{-}}\left[\mathcal{L}_{lg}\left(-f(\left<\mathcal{V}(s), \mathcal{V}(g)\right>)^{\top}f(\left<s,g\right>^{-})\right)\right]\\
&\leq\mathbb{E}_{\left<s,g\right>^{-}}\sup_{\mathcal{V}(s), \mathcal{V}(g)}\left[\mathcal{L}_{lg}\left(-f(\left<\mathcal{V}(s), \mathcal{V}(g)\right>)^{\top}f(\left<s,g\right>^{-})\right)\right].
\end{aligned}$}
\end{equation}
Consequently, we can reframe the challenge of computing the supremum over expectations into determining the expectation over the supremum. This means our goal is to optimize $\mathcal{L}_{lg}$ for a given negative tuple $\left<s, g\right>^{-}$. Moreover, leveraging the non-increasing nature of $\mathcal{L}_{lg}(\cdot)$, we can approximate our attack objective by directly minimizing the similarity-based loss $-f(\left<\mathcal{V}(s), \mathcal{V}(g)\right>)^{\top}f(\left<s,g\right>^{-})$. In this paper, $f(\cdot)$ is equivalent to $\psi(\cdot)$. Building on this, we present a Projected Gradient Descent (PGD)-based approximation for both $\mathcal{V}(s)$ and $\mathcal{V}(g)$.
\begin{Definition}[PGD-based Approximation]
\label{eas}
Given an encoder $\psi(\cdot)$, an original input tuple $\left<s,g\right>$, a pre-defined negative tuple $\left<s,g\right>^{-}$, and the step size $\alpha$, then the semi-contrastive representation attack $\mathcal{V}_{scr}(s)$ and $\mathcal{V}_{scr}(g)$ at the iterative step $i+1$ can be individually defined as:
{
\begin{equation}
\label{eq:pgd-based-approximation-1}
\mathcal{V}^{i+1}_{scr}(s) = \mathcal{V}^i_{scr}(s) -\alpha\nabla_{\mathcal{V}_{scr}^{i}(s)}\mathcal{L}_{sim}(s,g,i),
\end{equation}
\begin{equation}
\label{eq:pgd-based-approximation-2}
\mathcal{V}^{i+1}_{scr}(g) = \mathcal{V}^i_{scr}(g) -\alpha\nabla_{\mathcal{V}_{scr}^{i}(g)}\mathcal{L}_{sim}(s, g, i),
\end{equation}
}
Particularly, we define:
\begin{equation}
\resizebox{0.88\linewidth}{!}{
$\begin{aligned}
\mathcal{L}_{sim}(s, g, i) = -\psi\left(\left<\mathcal{V}_{scr}^{i}(s), \mathcal{V}^{i}_{scr}(g)\right>\right)^\top\psi(\left<s,g\right>^{-}).
\end{aligned}$}
\end{equation}
If the number of iteration steps is $\mathcal{I}(\geq$1$)$, then the finally generated adversarial tuple can be defined as $\left<\mathrm{proj}(\mathcal{V}^{I}_{scr}(s)), \mathrm{proj}(\mathcal{V}^{I}_{scr}(g))\right>$, where $\mathrm{proj}(\cdot)$ is the projection head. 
\end{Definition}
In this work, We denote $proj(\cdot)$ in Def.~\ref{eas} as an $\epsilon$-bounded $\ell_{\infty}$-norm ball. In a straightforward yet effective manner, we construct the negative tuple $\left<s,g\right>^{-}$ by performing negation operation, including $\left<-s, -g\right>$, $\left<-s, g\right>$ and $\left<s, -g\right>$. By incorporating the SCR attack at each timestep within an episode of $T$ steps in GCRL, our primary aim is to divert the agent, ensuring that it remains distant from the goal. This approach ensures that the reward sequence $(r_0, \cdots, r_{T-1})$ remains as sparse as possible.

To verify the efficacy of our approach, we train multiple agents using diverse GCRL algorithms and evaluate their adversarial robustness against our introduced attack. Comprehensive results are presented in the \textit{Experiments} section.

\section{Adversarial Representation Tactics}

As illustrated in Tab.~\ref{table:adversarial_attack} and \ref{table:defensive_simsr}, GCRL agents trained using both base methods and SimSR-strengthened ones are vulnerable to our SCR attack, highlighting a significant security concern. To address this, we introduce a composite defensive strategy termed ARTs in this section, which strategically combine the Semi-Contrastive Adversarial Augmentation (SCAA) and the Sensitivity-Aware Regularizer (SAR), catering to diverse GCRL algorithms.

\subsection{Semi-Contrastive Adversarial Augmentation}
In Algorithm~\ref{algorithm_scaa}, we investigate the influence of data augmentations by subjecting input tuples to the SCR attack. Specifically, during each training epoch for the base GCRL agent, we retrieve a mini-batch of tuples from the replay buffer and generate semi-contrastive augmented samples, $\mathcal{V}_{scr}(s_t^i)$ and $\mathcal{V}_{scr}(g^i)$, using the constructed negative tuples. We then utilize the augmented critic value, denoted as $\hat{\varrho}$, and the actor value, represented as $\hat{\varphi}$, to refine the weights of the encoder ($\theta_{\psi}$), actor network ($\theta_{\varphi}$), and critic network ($\theta_{\varrho}$). Notably, $\mathcal{L}_{\psi}$, $\mathcal{L}_{\varphi}$, and $\mathcal{L}_{\varrho}$ correspond to the loss functions for the encoder, actor network, and critic network, respectively.

\subsection{Sensitivity-Aware Regularizer}
Given two pairs of observations $\left(s_i, g_1, r_{\left<s_i, g_1\right>}, a_i, s_{i+1}\right)$ 
and $\left(s_j, g_{2}, r_{\left<s_j, g_2\right>}, a_j, s_{j+1}\right)$ shown in Fig.~\ref{figure:special_case}, it is common that $r_{\left<s_i, g_1\right>}=r_{\left<s_j, g_2\right>}=0$, due to the sparsity of reward sequences. Therefore, given the above pair of transitions, we can draw from Thm.~\ref{simsr}, and iteratively update the encoder $\psi(\cdot)$ in the policy function using the mean square loss:
\begin{equation}
\label{mse}
\resizebox{0.88\linewidth}{!}{
$\begin{aligned}
\left(\mathcal{M}(\left<s_i, g_1\right>,\left<s_j, g_2\right>)-\gamma\mathbb{E}_{s_{i+1}, s_{j+1}}\left[\mathcal{M}(\left<s_{i+1}, g_1\right>, \left<s_{j+1}, g_2\right>)\right]\right)^2,
\end{aligned}$}
\end{equation}
where $\mathcal{M}(\left<s_i, g_1\right>, \left<s_j, g_2\right>)$ is highly dependent on the next states $s_{i+1}$ and $s_{j+1}$, and rarely captures information from the absolute reward difference $\left|r_{\left<s_{i}, g_1\right>}-r_{\left<s_j, g_2\right>}\right|$. This cannot provide any evaluations of the difference between state-goal tuples at the current step, and prevents the policy and critic functions from gaining insight from the interaction between the agent and the environment. To compensate for this deficiency of SimSR utilized in GCRL, we construct the Sensitivity-Aware Regularizer as a substitute for $\left|r_{\left<s_i, g_1\right>}-r_{\left<s_j, g_2\right>}\right|$.
\begin{Definition}[Sensitivity-Aware Regularizer]
\label{sar}
Given perturbations $\delta_{s_i}$, $\delta_{g_1}$, $\delta_{s_j}$, $\delta_{g_2}$ 
for tuples $\left<s_i, g_1\right>$ and $\left<s_j, g_2\right>$, and a trade-off factor $\beta$, the sensitivity-aware regularizer for the encoder $\psi(\cdot)$ can be defined as:
\begin{equation}
\label{regularizer_sar}
\resizebox{0.88\linewidth}{!}{
$\begin{aligned}
&\left|\frac{\mathcal{M}(\left<s_i, g_1\right>, \left<s_i+\delta_{s_i},g_1\right>)}{\left\|\delta_{s_i}\right\|_2} -\frac{\mathcal{M}(\left<s_j,g_2\right>, \left<s_{j}+\delta_{s_j}, g_2\right>)}{\left\|\delta_{s_j}\right\|_2}\right|+\\
&\beta\left|\frac{\mathcal{M}(\left<s_i,g_1\right>, \left<s_{i}, g_1+\delta_{g_1}\right>)}{\left\|\delta_{g_1}\right\|_2} -\frac{\mathcal{M}(\left<s_j,g_2\right>, \left<s_{j}, g_2+\delta_{g_2}\right>)}{\left\|\delta_{g_2}\right\|_2}\right|,
\end{aligned}$}
\end{equation}
then we can empirically optimize the encoder parameters $\theta_{\psi}$ using the loss function combining Eq.~(\ref{mse}) and Eq.~(\ref{regularizer_sar}).
\end{Definition}
As shown in Def.~\ref{sar}, we explore the Lipschitz constant-based robustness term $\mathcal{M}(\cdot)/\left\|\delta\right\|_2$ to indicate the implicit difference. Generally, on the one hand, physically close tuples should have similar Lipschitz constants both on states and goals with a high probability, physically distant tuples. On the other hand, physically distant tuples tend to have dissimilar Lipschitz constants with a high probability. This reflects the similar characteristic as the absolute reward difference utilized in Eq.~(\ref{simsr_operator}). 

\input{Contents/algorithm}

%% file: Contents/algorithm.tex
\begin{algorithm}[tb]
\caption{Semi-Contrastive Adversarial Augmentation}
\textbf{Require:} Offline dataset $\mathcal{D}$,  learning rate $\alpha$, training steps $T$\\
\textbf{Initialisation:} Critic function $\varrho(\cdot)$, policy function $\varphi(\psi(\cdot))$
\begin{algorithmic}[1]
\For {training timestep $1\dots T$}
\State Sample a mini-batch from the offline dataset:

$(s_t^i, a_t^i, s_{t+1}^i, g^i) \sim \mathcal{D}$
\State Construct negative tuples:

$\left<s_t^i,g^i\right>^{-}$ and $\left<s_{t+1}^i,g^i\right>^{-}$
\State Compute the semi-contrastive augmented samples:

$\mathcal{V}_{scr}(s_t^i)$ and $\mathcal{V}_{scr}(g^i)$ by Eq.~(\ref{eq:pgd-based-approximation-1}-\ref{eq:pgd-based-approximation-2}) 
\State 
    Augment critic value:
    
    $\hat{\varrho} \gets \mathbb{E} [\varrho(s_{t}^i,g^i,a_t^i) + \varrho(\mathcal{V}_{scr}(s_{t}^i),\mathcal{V}_{scr}(g^i),a_t^i)]$
\State Augment actor value:

$\scriptstyle\hat{\varphi} \gets \mathbb{E}[\varphi(\psi(\left<s_t^i, g^i\right>)) + \varphi(\psi(\left<\mathcal{V}_{scr}(s_{t}^i), \mathcal{V}_{scr}(g^i)\right>)]$
\State Update encoder weights: 
$\scriptstyle\theta_{\psi} \gets \theta_\psi - \alpha \nabla \mathcal{L}_\psi(\left<s_t^i,g^i\right>)$
\State Update actor weights:
$\theta_\varphi \gets \theta_\varphi - \alpha \nabla \mathcal{L}_\varphi(\hat{\varphi})$
\State Update critic weights: 
$\theta_\varrho\gets \theta_\varrho-\alpha  \nabla \mathcal{L}_\varrho(\hat{\varrho})$
\EndFor
\end{algorithmic}
\label{algorithm_scaa}
\end{algorithm}

%% file: Contents/experiments.tex

\begin{table*}[htbp]
\begin{center}
\resizebox{\linewidth}{!}{\input{Tables/table_DisReturn_layer-3_epsilon_0.1_all}}
\end{center}
\caption{Comparison of discounted returns for DDPG, GCSL, and GoFar against various attack methods in GCRL. {\color[HTML]{FE0000}\textbf{Red}, \color[HTML]{6200C9}\textbf{Purple}}, and {\color[HTML]{3531FF}\textbf{Blue}} individually represent the worst, the second worst, and the third worst returns in each row. The experiments are averaged over 5 seeds.}
\label{table:adversarial_attack}
\end{table*}

\begin{table*}[htbp]
\resizebox{\linewidth}{!}
{\input{Tables/table_DisReturn_layer-3_epsilon_0.1_gofar+simsr_ddpg+simsr}}
\caption{Comparison of discounted returns for DDPG (SimSR) and GoFar (SimSR) against various attack methods in GCRL, averaged over 5 seeds. \textbf{Bold} denotes the worst return in each row.}
\label{table:defensive_simsr}
\end{table*}

\section{Experiments}

\begin{figure*}[tb] 
\centering
\includegraphics[width=\textwidth]{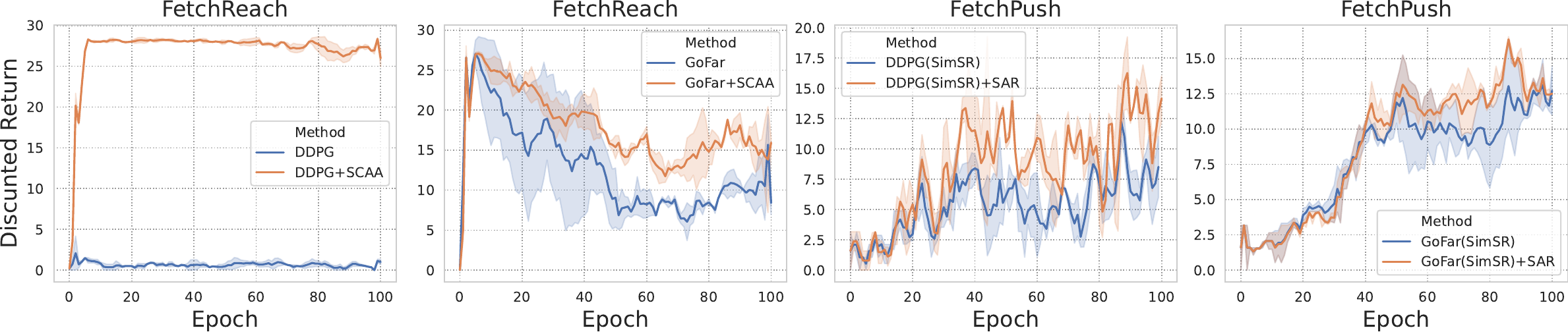}
\caption{Epoch-wise evaluations on the SCR attack of ARTs-defended DDPG, DDPG (SimSR), GoFar, and GoFar (SimSR).}
\label{figure:robust_training}
\end{figure*}

Our experimental design proceeds as follows: Initially, we identify the optimal representation layer by executing SCR attacks across various layers. Next, we benchmark the SCR attack against other adversarial attacks, using state-of-the-art algorithms as base methods. Finally, we achieve enhanced robustness of different base methods using ARTs.

\subsection{Baselines}
We evaluate our method\footnote{Our code is available at \url{github.com/TrustAI/ReRoGCRL}} using four robot manipulation tasks, namely FetchPush, FetchReach, FetchSlide, and FetchPick, as described in \cite{plappert2018multi}. The offline dataset for each task is collected through either a purely random policy or a combination of $90\%$ random policies and $10\%$ expert policies, depending on whether the random data sufficiently represents the desired goal distribution. We select 3
algorithms as our baselines for GCRL: Deep Deterministic Policy Gradient (DDPG) \cite{andrychowicz2018hindsight}, Goal-Conditioned Supervised Learning (GCSL) \cite{ghosh2020learning}, and Goal-Conditioned F-Advantage Regression (GoFar) \cite{ma2022offline}. It is worth noting that to circumvent costly and potentially risky environment interactions, we learn general goal-reaching policies from offline interaction datasets.

\subsection{Network Architectures}
\begin{figure}[tb] 
\centering
\subfigure{\includegraphics[width=0.99\linewidth]{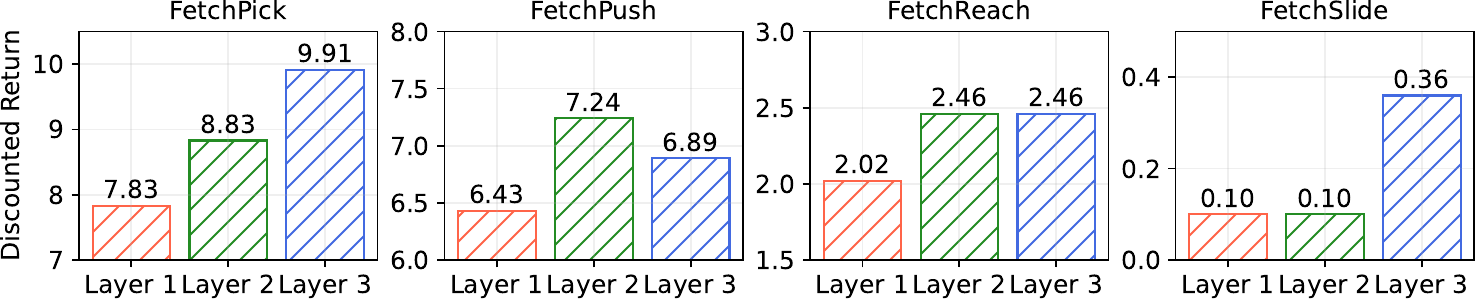}}
\subfigure{\includegraphics[width=0.99\linewidth]{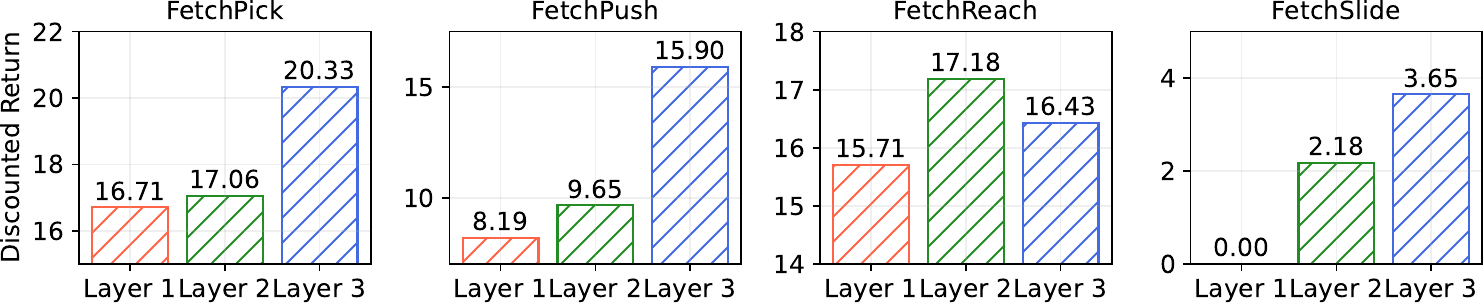}}
\caption{Evaluation of the SCR attack across various layers in MLP-based architectures. The \textbf{upper} charts use the DDPG method. The \textbf{lower} charts use the GoFar method.}
\label{figure:gcsl_layer}
\end{figure}

We use 3-layer MLPs as the backbones of $\varphi(\psi(\cdot))$ and $\varrho(\cdot)$. To determine $L_{\psi}$ and $L_{\varphi}$, we conduct preliminary experiments based on $\varphi(\psi(\cdot))$, trained via DDPG and GoFar to ascertain the optimal layer for representation. As depicted in Fig.~\ref{figure:gcsl_layer}, we compute the discounted returns across three distinct layers (Layer 1 denotes the first layer of $\varphi(\psi(\cdot))$, and so forth), with each layer acting as the representation layer. Comprehensive results indicate that Layer 1 is the most susceptible. Therefore, all attacks and training algorithms in this paper are computed on Layer 1.

In particular, the encoder $\psi(\cdot)$ processes the input concatenating a state $s$ and a goal $g$, which is represented as $o_0^{\pi}$. The dimension of the input tuple is $D_{sg}$. Thus, we define the latent representation given by $\psi(o_{0}^{\pi})=\phi(\mathbf{W}_1o_0^{\pi})$, $\mathbf{W}^{\pi}_1\in\mathbb{R}^{256\times D_{sg}}$. Subsequently, the actor network is constructed as $\varphi(\psi(o_0^{\pi}))=\mathbf{W}^{\pi}_4\phi(\mathbf{W}^{\pi}_3\phi(\mathbf{W}^{\pi}_2\psi(o_0^{\pi})))$, with $\mathbf{W}^{\pi}_2,\mathbf{W}^{\pi}_3 \in\mathbb{R}^{256\times256}$, $\mathbf{W}^{\pi}_4\in\mathbb{R}^{D_a\times256}$, where $D_a$ is the dimension of the action. Similar to the policy function $\pi$, we construct the output of critic function as $o_4^c=\mathbf{W}^{c}_4\phi(\mathbf{W}^{c}_3\phi(\mathbf{W}^{c}_2\phi(\mathbf{W}^{c}_1o^{c}_0)))$, where $o_0^c$ is the concatenation of the tuple $\left<s,g\right>$ and the action $a$, which has $(D_{sg}+D_{a})$ dimensions. Specifically, $\mathbf{W}_1^c\in\mathbb{R}^{256\times (D_{sg}+D_{a})}$,$\mathbf{W}_2^c$, $\mathbf{W}_3^c$ $\in\mathbb{R}^{256\times 256}$, and $\mathbf{W}_4^{c}\in\mathbb{R}^{1\times256}$.

\subsection{Comparison of SCR and Other Attacks}
Following Def.~\ref{eas}, we evaluate the robustness of multiple GCRL algorithms using $5$ different attack methods, including Uniform, SA-FGSM, SA-PGD, SCR-FGSM, and SCR-PGD. 
Specifically, we employ a $10$-step PGD with a designated step size of $0.01$. For both FGSM and PGD attacks, the attack radius is set to $0.1$.

Our SCR attack introduces perturbations in the form of $\left<\mathcal{V}(s), g\right>$, $\left<s, \mathcal{V}(g)\right>$, and $\left<\mathcal{V}(s), \mathcal{V}(g)\right>$, respectively. Specifically, Tab.~\ref{table:adversarial_attack}-\ref{table:defensive_simsr} presents the results when noise is added solely to the states. We provide the remaining attack results in the Appendix. As outlined in Def.~\ref{eas}, The columns \textit{state}, \textit{goal}, and \textit{state+goal} denote negative tuples as $\left<-s, g\right>$, $\left<s, -g\right>$, and $\left<-s, -g\right>$ individually.

As shown in Table~\ref{table:adversarial_attack}, for DDPG-based GCRL, we achieve $47.17\%$, $57.69\%$, $93.78\%$, and $100.00\%$ decrease of discounted returns in FetchPick, FetchPush, FetchReach, and FetchSlide respectively, outperforming SA-based attacks by $60.07\%$, $64.85\%$, $90.76\%$, and $100.00\%$. In detail, 
all of them are derived from the negative tuple $\left<-s, -g\right>$. The results are similar in GCSL, compared with the nature return, our method degrades the performance in FetchPick, FetchPush, and FetchSlide each by $18.41\%$, $35.68\%$, and $100.00\%$. Note that GCSL-based GCRL is more robust in FetchReach than DDPG-based one, our attack can only decrease the performance by $4.46\%$. As the state-of-the-art algorithm in GCRL, GoFar achieves better nature returns than DDPG and GCSL, thus all attacks exhibit a degradation in their attack capabilities. Particularly, our method achieves a reduction in overall returns of $17.28\%$, $40.04\%$, $58.71\%$, and $100.00\%$ across the four tasks, respectively. Similar to DDPG, the best-performing attack for each of these 4 tasks originates from $\left<-s, -g\right>$. 

We further evaluate the adversarial robustness of SimSR-enhanced algorithms as presented in Tab.~\ref{table:defensive_simsr}. 
All attack strategies employ the negative tuple $\left<-s, g\right>$, notably, our attack surpasses most other attacks when applied to DDPG.
However, when it comes to GoFar, our strategy faces more robust defensive measures. 



\subsection{Effectiveness of ARTs}
To test the efficacy of ARTs, we correspondingly perform attacks on ARTs-enhanced versions of different GCRL algorithms. Following Alg.~\ref{algorithm_scaa}, we train the agent using the SCR attack. For SimSR-boosted representations, we adaptively employ SAR during the optimization of $\psi(\cdot)$. Tab.~\ref{table:contrastive_robust_performance_main} only showcases the best attack results from Uniform, SA-FGSM, SA-PGD, SCR-FGSM, and SCR-PGD.  All attacks and training procedures in this section are based on  $\left<-s, g\right>$, with perturbations applied solely to the state positions.

As illustrated in Tab.~\ref{table:contrastive_robust_performance_main}, our defensive strategy significantly bolsters the robust performance of DDPG. Specifically, we register performance enhancements of $85.57\%$, $99.08\%$, and $1388.17\%$ in FetchPick, FetchPush and FetchReach, respectively. In a similar vein, for DDPG (SimSR), the implementation of ARTs leads to robustness improvements of $6.61\%$, $20.17\%$, $10.13\%$, and $48.65\%$ across FetchPick, FetchPush, FetchReach, and FetchSlide, respectively.  

For GoFar, our results are comparable and even superior in FetchPush, FetchReach, and FetchSlide. Additionally, for GoFar (SimSR), we observe a performance boost of $7.39\%$ and $149.15\%$, respectively, in FetchPick and FetchSlide. Complete results are available in the Appendix.

Qualitatively, Fig.~\ref{figure:robust_training} provides a visual representation of the epoch-wise performance of ARTs-enhanced GCRL algorithms in comparison to several base methods. In the FetchReach environment, both DDPG+SCAA and GoFar+SCAA exhibit notable improvements within the initial 100 epochs. A similar trend is observed in the FetchPush environment.

\begin{table}[tb]
\begin{center}
\resizebox{\linewidth}{!}{\input{Tables/DisReturn_layer-3_epsilon_0.1_all_Training_final}}
\end{center}
\caption{Defensive performance of ARTs on DDPG, GoFar, DDPG (SimSR), and GoFar (SimSR) against best attacks, averaged over 5 seeds. \textbf{Bold} indicates the better result in a block.}
\label{table:contrastive_robust_performance_main}
\end{table}



\label{experiments}

%% file: Tables/table_DisReturn_layer-3_epsilon_0.1_all.tex
{
\begin{tabular}{@{}c|c|c|ccc|ccc|ccc@{}}
\toprule
\multirow{3}{*}{Task}        & 
\multirow{3}{*}{Method} &  & \multicolumn{9}{c}{Attack Return}                      \\
                            &                                               & Nature & Uniform & SA-FGSM & SA-PGD & \multicolumn{3}{c}{SCR-FGSM} & \multicolumn{3}{|c}{SCR-PGD} \\ 
          & & & & &   & state & goal & state+goal & state & goal & state+goal \\            
                            \midrule
\multirow{3}{*}{FetchPick}  
& DDPG                    &14.82\tiny{$\pm$1.53} &16.68\tiny{$\pm$4.33} &19.61\tiny{$\pm$2.84} &19.92\tiny{$\pm$2.32} &11.18\tiny{$\pm$2.74}
&15.00\tiny{$\pm$1.88}  &\color[HTML]{FE0000} \textbf{7.83\tiny{$\pm$3.06}} &\color[HTML]{3531FF} \textbf{10.97\tiny{$\pm$4.12}} &16.38\tiny{$\pm$2.72} &\color[HTML]{6200C9} \textbf{10.49\tiny{$\pm$2.86}}      \\
& GCSL                    &11.39\tiny{$\pm$1.97}        &9.95\tiny{$\pm$1.65}              &9.94\tiny{$\pm$1.80}         &10.07\tiny{$\pm$}1.15        &9.52\tiny{$\pm$}2.13        &10.41\tiny{$\pm$}1.95 &\color[HTML]{3531FF} \textbf{8.99\tiny{$\pm$}1.56} &\color[HTML]{6200C9} \textbf{8.21\tiny{$\pm$3.12}} &11.72\tiny{$\pm$1.81} &\color[HTML]{FE0000} \textbf{8.11\tiny{$\pm$2.04}}      \\
& GoFar
&21.01\tiny{$\pm$}2.05
&19.91\tiny{$\pm$}2.39
&20.07\tiny{$\pm$}2.13
&18.92\tiny{$\pm$}1.80
&\color[HTML]{3531FF} \textbf{17.14\tiny{$\pm$}2.89}
&20.22\tiny{$\pm$}3.25 &\color[HTML]{6200C9} \textbf{16.71\tiny{$\pm$}3.26} &17.93\tiny{$\pm$}2.48
&19.84\tiny{$\pm$}1.88
&\color[HTML]{FE0000} \textbf{15.65\tiny{$\pm$}1.83}
                            \\ 
                            \midrule
\multirow{3}{*}{FetchPush} 
& DDPG                   &12.81\tiny{$\pm$4.61} &13.61\tiny{$\pm$4.81} &17.24\tiny{$\pm$2.12} &15.42\tiny{$\pm$3.38} &\color[HTML]{3531FF} \textbf{8.71\tiny{$\pm$3.22}} &10.27\tiny{$\pm$3.85} &\color[HTML]{6200C9} \textbf{6.43\tiny{$\pm$2.33}} &9.79\tiny{$\pm$2.15} &13.61\tiny{$\pm$2.56} &\color[HTML]{FE0000} \textbf{5.42\tiny{$\pm$3.00}}      \\

                                            & GCSL                    & 12.74\tiny{$\pm$1.38}       & 12.57\tiny{$\pm$1.67}         &9.64\tiny{$\pm$}3.28         & 9.71\tiny{$\pm$}3.37       &9.64\tiny{$\pm$}0.51        &12.97\tiny{$\pm$}3.09 &\color[HTML]{3531FF} \textbf{8.99\tiny{$\pm$}1.56} &\color[HTML]{FE0000} \textbf{6.20\tiny{$\pm$}3.66}  &13.56\tiny{$\pm$}1.49 &\color[HTML]{6200C9} \textbf{8.11\tiny{$\pm$}2.04}    \\
& GoFar
                            &18.38\tiny{$\pm$}3.24 
                            &16.11\tiny{$\pm$}2.39
                            &15.45\tiny{$\pm$}2.71
                            &13.66\tiny{$\pm$}3.97
                            &\color[HTML]{3531FF} \textbf{11.66\tiny{$\pm$}3.93}
                            &17.47\tiny{$\pm$}2.07 &\color[HTML]{FE0000} \textbf{8.19}\tiny{$\pm$}3.99 &12.14\tiny{$\pm$}3.78
                            &18.38\tiny{$\pm$}2.91
                            &\color[HTML]{6200C9} \textbf{10.57\tiny{$\pm$}3.93}\\ 
                            \midrule
\multirow{3}{*}{FetchReach}  
& DDPG                    &29.92\tiny{$\pm$0.20}      &28.86\tiny{$\pm$0.46}      &24.21\tiny{$\pm$2.21}       &20.14\tiny{$\pm$5.04}       &6.10\tiny{$\pm$7.33} &\color[HTML]{3531FF} \textbf{5.14\tiny{$\pm$7.19}} &\color[HTML]{6200C9} \textbf{2.02\tiny{$\pm$3.32}}  &7.02\tiny{$\pm$2.20} &11.03\tiny{$\pm$8.47} &\color[HTML]{FE0000} \textbf{1.86\tiny{$\pm$3.31}}    \\

                       & GCSL                    &22.04\tiny{$\pm$0.68}        &22.75\tiny{$\pm$0.82}               &22.01\tiny{$\pm$1.47}         &\color[HTML]{3531FF} \textbf{21.95\tiny{$\pm$1.51}}        &\color[HTML]{FE0000} \textbf{20.97\tiny{$\pm$1.31}}       &22.24\tiny{$\pm$0.78} &23.04\tiny{$\pm$1.62} &\color[HTML]{6200C9} \textbf{21.74\tiny{$\pm$0.81}} &22.22\tiny{$\pm$0.79}
                            &23.46\tiny{$\pm$1.14}\\
                                                & GoFar
                            
                            & 27.84\tiny{$\pm$0.63}
                            &27.57\tiny{$\pm$0.92}
                            &27.22\tiny{$\pm$0.49}
                            &27.22\tiny{$\pm$0.49}
                            &27.51\tiny{$\pm$0.45}
                            &27.45\tiny{$\pm$1.11} &15.71\tiny{$\pm$4.48} &27.96\tiny{$\pm$0.38} &27.87\tiny{$\pm$0.33} &\color[HTML]{FE0000} \textbf{11.24\tiny{$\pm$6.76}}
                            \\ 
                            \midrule
\multirow{3}{*}{FetchSlide} 
& DDPG                    &0.58\tiny{$\pm$1.30} &1.89\tiny{$\pm$0.89} &0.59\tiny{$\pm$1.09} &0.50\tiny{$\pm$1.12} &\color[HTML]{FE0000}\textbf{0.00\tiny{$\pm$0.00}} &\color[HTML]{6200C9} \textbf{0.36\tiny{$\pm$0.72}} &\color[HTML]{FE0000} \textbf{0.00\tiny{$\pm$0.00}} &\color[HTML]{FE0000} \textbf{0.00\tiny{$\pm$0.00}} &\color[HTML]{3531FF} \textbf{0.38\tiny{$\pm$0.60}} &\color[HTML]{FE0000} \textbf{0.00}\tiny{$\pm$0.00}\\

                                              & GCSL                    &1.55\tiny{$\pm$0.84}        &1.84\tiny{$\pm$1.19}                  &0.60\tiny{$\pm$0.42}         &0.57\tiny{$\pm$0.59}        &\color[HTML]{6200C9} \textbf{0.28\tiny{$\pm$0.63}}        &0.97\tiny{$\pm$0.63} &\color[HTML]{3531FF} \textbf{0.37\tiny{$\pm$0.51}} &\color[HTML]{3531FF} \textbf{0.37\tiny{$\pm$0.84}} &0.96\tiny{$\pm$1.26} &\color[HTML]{FE0000} \textbf{0.00}\tiny{$\pm$0.00}     \\
                                        & GoFar
                            &2.55\tiny{$\pm$1.37}
                            &1.23\tiny{$\pm$1.14}
                            &\color[HTML]{3531FF} \textbf{0.77\tiny{$\pm$0.92}}
                            &1.13\tiny{$\pm$0.72}
                            &\color[HTML]{6200C9} \textbf{0.09\tiny{$\pm$0.21}}
                            &1.25\tiny{$\pm$0.55} &\color[HTML]{FE0000} \textbf{0.00}\tiny{$\pm$0.00} 
                            &\color[HTML]{FE0000} \textbf{0.00\tiny{$\pm$0.00}}
                            &1.86\tiny{$\pm$0.73} &\color[HTML]{6200C9} \textbf{0.09\tiny{$\pm$0.22}}
                            \\ 
                            \bottomrule
\end{tabular}
}

%% file: Tables/table_DisReturn_layer-3_epsilon_0.1_gofar+simsr_ddpg+simsr.tex
{
\begin{tabular}{@{}c|cccccc@{}}
\toprule
&   \multicolumn{6}{c}{DDPG (SimSR)}         \\     
\multirow{-2}{*}{Task}         
 & Nature & Uniform& SA-FGSM & SA-PGD  & SCR-FGSM & SCR-PGD \\ \midrule
\multicolumn{1}{l|}{FetchPick}  &16.78 &16.00 &19.51 &17.92 & \textbf{15.74} &16.15 \\
\multicolumn{1}{l|}{FetchPush}  &12.63 &15.43 &16.24 &14.61 &10.70 & \textbf{10.51}\\
\multicolumn{1}{l|}{FetchReach}  &29.90 &29.46 &28.68 & \textbf{26.76} &27.72 &27.16 \\
\multicolumn{1}{l|}{FetchSlide} &0.62 &0.83 &0.80 &0.43 & \textbf{0.37} &0.84 \\ 
\bottomrule              
\end{tabular}

\begin{tabular}{@{}c|cccccc@{}}
\toprule
&   \multicolumn{6}{c}{GoFar (SimSR)}         \\           
 & Nature & Uniform& SA-FGSM & SA-PGD  & SCR-FGSM & SCR-PGD \\ \midrule
 &17.44 &16.73 &17.63 & \textbf{13.80} &16.27 &16.11 \\
&14.19 &13.13 &15.52 &13.91 & \textbf{12.44} &13.08\\
 &27.93 & \textbf{27.91} &27.95 &27.96 &27.96 &27.98 \\
&1.66 &1.78 &2.31 &1.41 &1.07 & \textbf{0.59} \\ 
\bottomrule              
\end{tabular}
}

%% file: Tables/DisReturn_layer-3_epsilon_0.1_all_Training_final.tex
\begin{tabular}{@{}cc|c|c|c|c@{}}
\toprule
\multicolumn{2}{c|}{Methods}                                                                                      & FetchPick             & FetchPush             & FetchReach            & FetchSlide \\ \midrule
\multirow{4}{*}{\rotatebox[origin=c]{90}{DDPG}}  & \multicolumn{1}{c|}{Vanilla}                           &7.83\scriptsize{$\pm$3.06}  &5.42\scriptsize{$\pm$3.00}  &1.86\scriptsize{$\pm$3.31}  &0.00\scriptsize{$\pm$0.00}            \\
 
& \multicolumn{1}{c|}{Vanilla+SCAA}  &\textbf{14.53\scriptsize{$\pm$3.08}}
&\textbf{10.79\scriptsize{$\pm$4.55}} 
&\textbf{27.68\scriptsize{$\pm$1.30}}  &\textbf{0.00\scriptsize{$\pm$0.00}}            \\
\cmidrule(l){2-6}
& \multicolumn{1}{c|}{SimSR}                &15.74\scriptsize{$\pm$}0.65 
&10.51\scriptsize{$\pm$}5.79 
&27.16\scriptsize{$\pm$}2.75
&0.37\scriptsize{$\pm$}0.60 \\
& \multicolumn{1}{c|}{SimSR+SAR}
&\textbf{16.78\scriptsize{$\pm$}1.39}       &\textbf{12.63\scriptsize{$\pm$3.89}} 
&\textbf{29.91\scriptsize{$\pm$0.20}}
&\textbf{0.55\scriptsize{$\pm$0.20}}
                       \\ \midrule
\multirow{4}{*}{\rotatebox[origin=c]{90}{GoFar}} & \multicolumn{1}{c|}{Vanilla}                     &\textbf{15.65\scriptsize{$\pm$}1.83}
&\textbf{8.19\scriptsize{$\pm$}3.99}
&11.24\scriptsize{$\pm$6.76} 
&0.00\scriptsize{$\pm$0.00}
\\
& \multicolumn{1}{c|}{Vanilla+SCAA} &14.72\scriptsize{$\pm$4.40} &8.00\scriptsize{$\pm$}3.76
&\textbf{13.99\scriptsize{$\pm$6.46}} &\textbf{0.00\scriptsize{$\pm$0.00}}            \\
\cmidrule(l){2-6} 
& \multicolumn{1}{c|}{SimSR}                &13.80\scriptsize{$\pm$}4.40 &\textbf{12.44\scriptsize{$\pm$}3.80}  &\textbf{27.91\scriptsize{$\pm$}0.34} &0.59\scriptsize{$\pm$}0.66            \\
& \multicolumn{1}{c|}{SimSR+SAR}   
&\textbf{14.82\scriptsize{$\pm$}3.04}             &\textbf{12.44\scriptsize{$\pm$}3.80}
&27.84\scriptsize{$\pm$}0.32
&\textbf{1.47\scriptsize{$\pm$}0.79}            \\ \bottomrule
\end{tabular}

%% file: Contents/conclusion.tex
\section{Conclusion}
Due to the sparse rewards in GCRL, we introduce the Semi-Contrastive Representation attack to probe its vulnerability, which only requires access to the policy function. Accordingly, we propose Adversarial Representation Tactics to bolster the robust performance of the underlying agent. 
In particular, we devise the Semi-Contrastive Adversarial Augmentation for baseline methods in GCRL and introduce the Sensitivity-Aware Regularizer for their robust counterparts that are guided by the bisimulation metric.
We utilize SimSR to learn invariant representations. The effectiveness of our methods is validated by intensive empirical experiments.

%% file: Contents/acknowledgement.tex
\section{Acknowledgements}
This work is supported by the UK EPSRC under project EnnCORE [EP/T026995/1], the University of Liverpool and the China Scholarship Council.

%% file: Contents/appendix.tex
\clearpage

\onecolumn
\section{Appendix}
In this section, we offer additional experimental details that are not covered in the \textit{Experiments} section of the main manuscript. This encompasses:
\begin{itemize}
\item Implementation details for DDPG, GCSL, and GoFar in GCRL.
\item Hyperparameters of training algorithms in GCRL.
\item Additional SCR attack results for DDPG, GCSL, and GoFar.
\item Defensive performance of ARTs on DDPG, GoFar, DDPG (SimSR) and GoFar (SimSR) against SCR-FGSM and SCR-PGD.
\item Other definitions for SA and SCR attacks.
\end{itemize}

\subsection{Baseline Implementations}
\textbf{DDPG} 
We utilize an open-source implementation of DDPG that has been fine-tuned for the set of Fetch tasks. The objective of the critic network is:

\begin{equation}
\begin{aligned}
\underset{Q}{\mathrm{min}} \mathbb{E}_{(s_t,a_t,s_{t+1},g)\sim\mathcal{D}}[(r(s_t,g)+\gamma\bar{Q}(s_{t+},\pi(s_{t+1},g),g)-Q(s_t,a_t,g))^2],
\end{aligned}
\end{equation}

\noindent where $\bar{Q}$ is the targeted Q network. The objective of the policy network is: 
\begin{equation}
\begin{aligned}
\underset{\pi}{\mathrm{min}} -\mathbb{E}_{(s_t,a_t,s_{t+1},g)\sim\mathcal{D}}[Q(s_t,\pi(s_t,g),g)].
\end{aligned}
\end{equation}

\noindent\textbf{GCSL} GCSL is a straightforward supervised regression method that employs behaviour cloning from a hindsight relabelling dataset. We execute the implementation of GCSL by excluding the DDPG critic component and altering the policy loss to adhere to maximum likelihood.
\begin{equation}
\begin{aligned}
\underset{\pi}{\mathrm{min}} -\mathbb{E}_{(s_t,a_t,g)\sim\mathcal{D}}[Q(s_t,\pi(s_t,g),g)].
\end{aligned}
\end{equation}

\noindent\textbf{GoFar} GoFAR benefits from uninterleaved optimization for its value and policy networks, eliminating the need for hindsight relabeling. They use a discriminator to estimate the reward as below:
\begin{equation}
\begin{aligned}
\mathcal{L}_c(\psi)=\frac{1}{N}\sum_{i=1}^N[\mathrm{log}(1-c_{\psi}(s_d^j,g^j))+\frac{1}{M}\sum_{j=1}^M[\mathrm{log}c_{\psi}(s_d^j,g^j)]],
\end{aligned}
\end{equation}
The value function is trained as follows:
\begin{equation}
\begin{aligned}
\mathcal{L}_V(\theta)=\frac{1-\gamma}{M}\sum_{i=1}^M[V_\theta(s_0^i,g_0^i)]+\frac{1}{N}\sum_{i=1}^N[f_*(R_t^i+\gamma V(s_{t+1}^i,g_t^i)-V(s_{t}^i,g_t^i))],
\end{aligned}
\end{equation}
Once the optimal $V^*$ has been obtained, the policy is learned via:
\begin{equation}
\begin{aligned}
\mathcal{L}_{\pi}(\phi)=\frac{1}{N}[f_*^{\prime}(R_t^i+\gamma V_{\theta}(s_{t+1}^i,g_t^i)-V_{\theta}(s_{t}^i,g_t^i))\mathrm{log}\pi(a|s,g)],
\end{aligned}
\end{equation}
All hyperparameters are kept the same as the original work, and the detailed algorithm can be found in Algorithm 2 in \cite{ma2022offline}.

\subsection{Hyperparameters}
In this work, DDPG, GCSL, and GoFar share the same network
architectures.  For all experiments, We train each method for 5 seeds, and each training run uses 100 epochs, containing a minibatch with size 512. The hyperparameters of our backbone architectures and algorithms are reported in Tab. ~\ref{table:hyperparameter}.


\begin{table*}[h]
\centering
\resizebox{0.35\linewidth}{!}{\input{Tables/hyperparameter}}
\caption{Hyperparameters for base methods in GCRL.}
\label{table:hyperparameter}
\end{table*}

\begin{table*}
\begin{center}
\resizebox{1.0\linewidth}{!}{\input{Tables/attack_ablation}}
\end{center}
\caption{Comparison of discounted returns for DDPG, GCSL, and GoFar against SCR-FGSM and SCR-PGD. The left sub-table shows the attack results added on both $s$ and $g$, while the right sub-table shows the attack results added on $g$. The experiments are averaged over 5 random seeds.}
\label{table:attack_ablation}
\end{table*}

\subsection{Task Details}
Each of these 4 tasks has continuous state, action, and goal space. The maximum episode horizon is defined as 50. The environments are built upon the 7-degree-of-freedom (DoF) Fetch robotics arm, equipped with a two-fingered parallel gripper. In this section, we provide a detailed introduction to these tasks.

\textbf{FetchPick} In this task, the objective is to grasp an object and transport it to a designated target location, which is randomly selected within the space (e.g., on the table or above it). The goal is represented by three dimensions, specifying the desired object position.

\textbf{FetchPush} In this task, a block is positioned in front of the robot. The robot's actions are limited to pushing or rolling the object to a specified target location. The goal is described by three dimensions, indicating the desired position of the object. 
The state space encompasses 25 dimensions, encompassing attributes such as the gripper's position, and linear velocities, as well as the position, rotation, and linear and angular velocities of the box.

\textbf{FetchReach} Within this setting, a 7 DoF robotic arm is tasked with accurately reaching a designated location using its two-fingered gripper. The state space encompasses 10 dimensions, including attributes like the gripper's position and linear velocities. Correspondingly, the action space is composed of 4 dimensions, representing the gripper's movements as well as its open/close status.

\textbf{FetchSlide} The FetchSlide environment is similar to FetchPush, involving the control of a 7 DoF robotic arm to slide a box towards a designated location. The key distinction lies in the fact that the target position lies beyond the robot's immediate reach. Therefore, the robot is required to strike the block to initiate its sliding motion towards the target position.

\subsection{Additional Results of the SCR Attack}
Besides the previous attack results shown in Tab.~\ref{table:adversarial_attack} and Tab.~\ref{table:defensive_simsr}, we conduct extra experiments in this section. As illustrated in Tab.~\ref{table:attack_ablation}, we evaluate the attack results derived from perturbations on both states and goals and solely on goals.



\begin{table*}[tb]
\begin{center}
\resizebox{0.9\linewidth}{!}{\input{Tables/DisReturn_layer-3_epsilon_0.1_all_Training}}
\end{center}
\caption{Defensive performance of ARTs on DDPG, GoFar, DDPG (SimSR) and GoFar (SimSR) against Uniform, SA-FGSM, SA-PGD, SCR-FGSM, and SCR-PGD attacks. The experiments are averaged over 5 random seeds. \textbf{Bold} indicates the best attack result in a row.}
\label{table:contrastive_robust_performance}
\end{table*}

\subsection{Full Results for Defensive Performance of ARTs}
Building upon the results presented in Tab.~\ref{table:contrastive_robust_performance}, we provide a comprehensive analysis of the defensive performance of DDPG, GoFar, and their ARTs-enhanced counterparts. The experimental setup is the same as the \textit{Experiments} section.
Details are illustrated in Tab. ~\ref{table:contrastive_robust_performance}.

\textbf{FetchPick}: Even though all algorithms undergo training with perturbations introduced by the SCR attack, we observed a boost in defensive performance against SA-FGSM for DDPG and DDPG (SimSR). A similar trend is evident against SA-PGD for DDPG (SimSR) and GoFar (SimSR). When defending against SCR-FGSM, the performance improvements are 102.43\%, 6.60\%, 0.66\%, and 7.19\% for DDPG, DDPG (SimSR), GoFar, and GoFar (SimSR) respectively. For SCR-PGD, the increments were 38.51\%, 3.90\%, and 8.26\% for DDPG, DDPG (SimSR), and GoFar (SimSR) respectively.

\textbf{FetchPush}: We achieve the performance enhancements when defending against SA-FGSM for DDPG, GoFar, and GoFar (SimSR). Similarly, improvements are observed against SA-PGD for DDPG, GoFar, and GoFar (SimSR). In the context of SCR-FGSM, the increments are 67.81\%, 18.04\%, 0.85\%, and 14.07\% for DDPG, DDPG (SimSR), GoFar, and GoFar (SimSR) respectively. For SCR-PGD, the gains were 164.94\%, 20.17\%, and 8.49\% for DDPG, DDPG (SimSR), and GoFar (SimSR) respectively.

\textbf{FetchReach}: The defence against SA-FGSM sees improvements for DDPG, GoFar, and GoFar (SimSR). Similarly, the performance against SA-PGD also rises for DDPG, GoFar, and GoFar (SimSR). For SCR-FGSM, the gains are 1270.30\% and 7.90\% for DDPG and DDPG (SimSR), respectively. Against SCR-PGD, the increments are 1403.23\%, 10.13\%, and 24.47\% for DDPG, DDPG (SimSR), and GoFar, respectively.

\textbf{FetchSlide}: Defense improvements against SA-FGSM are observed for DDPG and GoFar, while for SA-PGD, the gains were seen for DDPG, DDPG (SimSR), and GoFar (SimSR). Remarkably, for SCR-FGSM, the improvements were 67.57\% and 55.14\% for DDPG (SimSR) and GoFar (SimSR) respectively. Against SCR-PGD, the gains were 181.36\% for GoFar (SimSR).

\subsection{Others}
To streamline the main text, we reserve some definitions in this section.

\begin{Definition}[FGSM-based Approximation]
\label{fgsm_sgo}
Given an input tuple $\left<s,g\right>$, attack radius $\epsilon$, and loss function $\mathcal{L}$, the FGSM-based adversarial state and goal can be individually denoted as:
\begin{align}
\begin{split}
&\mathcal{V}(s) = s+\epsilon\nabla_{s}\mathcal{L}(\left<s,g\right>)\\
&\mathcal{V}(g) = g+\epsilon\nabla_{g}\mathcal{L}(\left<s,g\right>)
\end{split}
\end{align}
\end{Definition}
Different from Def.~\ref{eas}, Def.~\ref{fgsm_sgo} provides a single-step attack on $\left<s,g\right>$, where $\mathcal{V}(s)$ generated by FGSM is equivalent to $\mathcal{V}^{\mathcal{I}}(s)$ generated by PGD, $\mathcal{V}(g)$ generated by FGSM is equivalent to $\mathcal{V}^{\mathcal{I}}(g)$ generated by PGD. Similar to Def.~\ref{eas}, $\mathcal{I}$ is the number of iteration steps.
\begin{Definition}[SA attack]
\label{sa_attack}
Generally, given an input tuple $\left<s,g\right>$ and an action $a$, the SA attack can be defined as follows:
\begin{equation}
\mathop{\arg\sup}_{\mathcal{V}(s)\in\mathcal{B}_p^{\epsilon}(s), \mathcal{V}(g)\in\mathcal{B}_p^{\epsilon}(g)}\left[-Q(\left<s, g\right>, a)\right]
\end{equation}
\end{Definition}
Then PGD and FGSM variants of Def.~\ref{sa_attack} can be analogously represented following Def.~\ref{eas} and Def.~\ref{fgsm_sgo}.

%% file: Tables/hyperparameter.tex
\begin{tabular}{@{}c|c|c@{}}
\toprule
& \textbf{Hyperparameter} & \textbf{Value}       \\ \midrule
\multirow{1}{*}{Training} & Optimizer               & Adam                 \\
& Policy learning rate    & \(1 \times 10^{-3}\) \\
& Actor learning rate     & \(1 \times 10^{-3}\) \\
& Number of epochs        & 100                  \\
& Number of cycles        & 20                   \\
& Buffer size             & 2$\times10^6$        \\
& Discount factor         & 0.98                 \\
& Number of Tests & 10 \\
& Critic hidden dim & 256 \\
& Critic hidden layers & 3 \\
& Critic activation function & ReLU \\
& Actor hidden dim & 256 \\
& Actor hidden layers & 3 \\
\bottomrule
\end{tabular}

%% file: Tables/attack_ablation.tex
{
\begin{tabular}{@{}c|c|ccc|ccc|ccc|ccc@{}}
\toprule
\multirow{3}{*}{Task}        & 
\multirow{3}{*}{Method} &  \multicolumn{6}{c|}{Add on $s$ and $g$} & \multicolumn{6}{c}{Add on $g$}                 \\
                            &                                   & \multicolumn{3}{c}{SCR-FGSM}  & \multicolumn{3}{c}{SCR-PGD} & \multicolumn{3}{|c}{SCR-FGSM} & \multicolumn{3}{c}{SCR-PGD} \\ 
           &    & state & goal & state+goal & state & goal & state+goal & state & goal & state+goal & state & goal & state+goal \\            
                            \midrule
\multirow{3}{*}{FetchPick}  
& DDPG                       
&11.18\tiny{$\pm$2.74}
&14.79\tiny{$\pm$1.39} 
&11.39\tiny{$\pm$3.41}
&10.84\tiny{$\pm$3.59}
&14.97\tiny{$\pm$1.56}
&10.16\tiny{$\pm$2.68}
&15.70\tiny{$\pm$3.87}
&16.94\tiny{$\pm$2.87}
&16.76\tiny{$\pm$2.65}
&15.11\tiny{$\pm$2.45}
&16.38\tiny{$\pm$2.72}
&13.31\tiny{$\pm$2.94}
\\
& GCSL                            
&8.06\tiny{$\pm$}2.64
&11.14\tiny{$\pm$}1.92
&6.10\tiny{$\pm$}2.43
&8.18\tiny{$\pm$}3.10
&11.30\tiny{$\pm$}1.93
&6.12\tiny{$\pm$}3.31
&10.85\tiny{$\pm$}1.23
&11.66\tiny{$\pm$}1.95
&9.73\tiny{$\pm$}2.09
&11.17\tiny{$\pm$}2.05
&11.72\tiny{$\pm$}1.81
&9.48\tiny{$\pm$}2.45
\\
& GoFar
&17.14\tiny{$\pm$}2.89
&20.98\tiny{$\pm$}2.06
&16.23\tiny{$\pm$}2.40
&17.83\tiny{$\pm$}2.06
&21.01\tiny{$\pm$}2.05
&14.77\tiny{$\pm$}2.45
&20.63\tiny{$\pm$}2.19
&19.78\tiny{$\pm$}1.94
&14.79\tiny{$\pm$}3.59
&21.03\tiny{$\pm$}2.03
&19.84\tiny{$\pm$}1.88
&15.40\tiny{$\pm$}2.91
                            \\ 
                            \midrule
\multirow{3}{*}{FetchPush} 
& DDPG                    &6.22\tiny{$\pm$2.06}
&11.50\tiny{$\pm$2.50} 
&4.64\tiny{$\pm$2.20}
&9.79\tiny{$\pm$2.15}
&12.80\tiny{$\pm$4.61}
&6.08\tiny{$\pm$4.04}
&12.72\tiny{$\pm$4.07}
&12.57\tiny{$\pm$4.19} 
&13.41\tiny{$\pm$2.40}
&11.72\tiny{$\pm$2.56}
&13.61\tiny{$\pm$2.56}
&8.10\tiny{$\pm$2.67}\\

& GCSL              
&5.86\tiny{$\pm$}1.53
&12.63\tiny{$\pm$}2.02
&5.48\tiny{$\pm$}2.90
&6.20\tiny{$\pm$}2.11
&12.73\tiny{$\pm$}1.39
&6.39\tiny{$\pm$}2.71
&12.95\tiny{$\pm$}2.06
&13.37\tiny{$\pm$}0.90
&11.36\tiny{$\pm$}3.01
&12.90\tiny{$\pm$}2.50
&13.56\tiny{$\pm$}1.49
&10.09\tiny{$\pm$}3.29
\\
& GoFar
&11.66\tiny{$\pm$}3.93
&18.32\tiny{$\pm$}3.77
&10.85\tiny{$\pm$}4.41
&14.55\tiny{$\pm$}2.78
&18.38\tiny{$\pm$}3.24
&11.81\tiny{$\pm$}3.67
&16.65\tiny{$\pm$}1.96
&16.33\tiny{$\pm$}3.17
&10.95\tiny{$\pm$}2.80
&18.34\tiny{$\pm$}3.99
&18.83\tiny{$\pm$}1.39
&10.62\tiny{$\pm$}1.60
                            \\ 
                            
                            \midrule
\multirow{3}{*}{FetchReach}  
& DDPG                            
&27.42\tiny{$\pm$1.32}
&28.28\tiny{$\pm$0.70}
&27.33\tiny{$\pm$1.43}
&27.02\tiny{$\pm$2.20}
&29.91\tiny{$\pm$0.20}
&24.70\tiny{$\pm$4.07} 
&28.48\tiny{$\pm$0.55}
&10.40\tiny{$\pm$8.64}
&21.28\tiny{$\pm$7.45}
&29.91\tiny{$\pm$0.20}
&11.03\tiny{$\pm$8.47}
&6.07\tiny{$\pm$5.32}\\

& GCSL                            
&20.97\tiny{$\pm$1.31}
&22.14\tiny{$\pm$0.76}
&20.60\tiny{$\pm$1.26}
&21.18\tiny{$\pm$0.95}
&22.04\tiny{$\pm$0.68}
&20.81\tiny{$\pm$0.95}
&22.94\tiny{$\pm$0.75}
&22.22\tiny{$\pm$0.79}
&20.60\tiny{$\pm$1.26} 
&22.04\tiny{$\pm$0.68}
&22.22\tiny{$\pm$0.79}
&20.81\tiny{$\pm$0.95}
\\
& GoFar
&27.51\tiny{$\pm$0.45}
&27.66\tiny{$\pm$0.95}
&25.91\tiny{$\pm$2.02}
&27.49\tiny{$\pm$0.94}
&27.84\tiny{$\pm$0.63}
&26.05\tiny{$\pm$1.85}
&27.65\tiny{$\pm$0.80}
&27.85\tiny{$\pm$0.34} 
&22.57\tiny{$\pm$4.15}
&27.83\tiny{$\pm$0.65}
&27.87\tiny{$\pm$0.33}  
&24.18\tiny{$\pm$2.43}
\\ 
                            \midrule
\multirow{3}{*}{FetchSlide} 
& DDPG                    
&0.00\tiny{$\pm$0.00}
&0.22\tiny{$\pm$0.32}
&0.00\tiny{$\pm$0.00}
&0.00\tiny{$\pm$0.00}
&0.56\tiny{$\pm$1.27}
&0.00\tiny{$\pm$0.00}
&0.91\tiny{$\pm$0.95}
&0.60\tiny{$\pm$0.55} 
&0.13\tiny{$\pm$0.31}
&0.08\tiny{$\pm$0.18}
&0.38\tiny{$\pm$0.60} 
&0.84\tiny{$\pm$0.68} \\

& GCSL         
&0.28\tiny{$\pm$0.63}
&1.57\tiny{$\pm$1.06}
&0.33\tiny{$\pm$0.74}
&0.00\tiny{$\pm$0.00}
&1.43\tiny{$\pm$0.62}
&0.39\tiny{$\pm$0.89}
&1.96\tiny{$\pm$0.98}
&0.87\tiny{$\pm$0.52}
&1.40\tiny{$\pm$1.38}
&1.33\tiny{$\pm$0.67}
&0.96\tiny{$\pm$1.26}
&0.62\tiny{$\pm$0.80}
\\
& GoFar
                
&0.09\tiny{$\pm$0.21}
&2.10\tiny{$\pm$1.38}
&0.25\tiny{$\pm$0.55}
&0.00\tiny{$\pm$0.00}
&2.55\tiny{$\pm$1.37}
&0.16\tiny{$\pm$0.22}
&2.36\tiny{$\pm$1.46}
&1.73\tiny{$\pm$0.74}
&1.18\tiny{$\pm$1.17}
&2.68\tiny{$\pm$1.01} 
&1.86\tiny{$\pm$0.73}
&0.64\tiny{$\pm$0.73}
                            \\ 
                            \bottomrule
\end{tabular}
}

%% file: Tables/DisReturn_layer-3_epsilon_0.1_all_Training.tex
\begin{tabular}{@{}c|c|c|c|ccccc@{}}
\toprule
\multirow{2}{*}{Env}        & \multirow{2}{*}{$\epsilon$} & \multirow{2}{*}{Method} & \multirow{2}{*}{Nature} & \multicolumn{5}{c}{Attack Return}                      \\
&                          & &  & Uniform  & SA-FGSM & SA-PGD & SCR-FGSM & SCR-PGD 
\\ \midrule
\multirow{8}{*}{FetchPick}  & \multirow{8}{*}{0.1}     & DDPG      &14.82\scriptsize{$\pm$1.53} &16.68\scriptsize{$\pm$4.33} &19.61\scriptsize{$\pm$2.84} &19.92\scriptsize{$\pm$2.32}  &\textbf{7.83}\scriptsize{$\pm$3.06} &10.49\scriptsize{$\pm$2.86}       \\
& & DDPG+SCAA &19.41\scriptsize{$\pm$2.59} &18.75\scriptsize{$\pm$2.65} &20.84\scriptsize{$\pm$1.47} &19.73\scriptsize{$\pm$1.93} &15.85\scriptsize{$\pm$2.65} &\textbf{14.53}\scriptsize{$\pm$3.08} \\
&
& DDPG(SimSR)  &16.78\scriptsize{$\pm$}1.39 &16.00\scriptsize{$\pm$}1.68 &19.51\scriptsize{$\pm$}1.05 &17.92\scriptsize{$\pm$}1.65 &\textbf{15.74}\scriptsize{$\pm$}0.65 &16.15\scriptsize{$\pm$}3.49 \\
& & DDPG(SimSR)+SAR &16.78\scriptsize{$\pm$}1.39 &18.16\scriptsize{$\pm$}0.87 &19.51\scriptsize{$\pm$}1.05
&18.92\scriptsize{$\pm$}0.94
&\textbf{16.78}\scriptsize{$\pm$}1.39
&\textbf{16.78}\scriptsize{$\pm$}1.39
\\
&    & GoFar
&21.01\scriptsize{$\pm$}2.05
&19.91\scriptsize{$\pm$}2.39
&20.07\scriptsize{$\pm$}2.13
&18.92\scriptsize{$\pm$}1.80
&16.71\scriptsize{$\pm$}3.26
&\textbf{15.65}\scriptsize{$\pm$}1.83
                            \\ 
&
& GoFar+SCAA &19.83\scriptsize{$\pm$3.94} &18.99\scriptsize{$\pm$3.57} &17.89\scriptsize{$\pm$4.13}
&16.24\scriptsize{$\pm$3.86}
&16.82\scriptsize{$\pm$3.46}
&\textbf{14.72}\scriptsize{$\pm$4.40}\\
& &GoFar(SimSR) &17.44\scriptsize{$\pm$}4.40 &16.73\scriptsize{$\pm$}3.81
&17.63\scriptsize{$\pm$}3.61 
&\textbf{13.80}\scriptsize{$\pm$}4.40&16.27\scriptsize{$\pm$}2.75 
&16.11\scriptsize{$\pm$}3.49 \\
& &GoFar(SimSR)+SAR 
&17.44\scriptsize{$\pm$}2.95
&17.41\scriptsize{$\pm$}3.37
&17.54\scriptsize{$\pm$}4.23
&\textbf{14.82}\scriptsize{$\pm$}3.04
&17.44\scriptsize{$\pm$}2.95
&17.44\scriptsize{$\pm$}2.95
\\
\midrule
\multirow{8}{*}{FetchPush} & \multirow{8}{*}{0.1}     & DDPG                   &12.81\scriptsize{$\pm$4.61} &13.61\scriptsize{$\pm$4.81} &17.24\scriptsize{$\pm$2.12} &15.42\scriptsize{$\pm$3.38}  &6.43\scriptsize{$\pm$2.33}  &\textbf{5.42}\scriptsize{$\pm$3.00}        \\

& &DDPG+SCAA &12.21\scriptsize{$\pm$4.16} &13.28\scriptsize{$\pm$3.72} &19.84\scriptsize{$\pm$2.50} &17.73\scriptsize{$\pm$2.77} &\textbf{10.79}\scriptsize{$\pm$4.55} &14.36\scriptsize{$\pm$2.90} \\
& &DDPG(SimSR) &12.63\scriptsize{$\pm$}3.89
&15.43\scriptsize{$\pm$}5.36 
&16.24\scriptsize{$\pm$}2.42 
&14.61\scriptsize{$\pm$}4.41
&10.70\scriptsize{$\pm$}4.55
&\textbf{10.51}\scriptsize{$\pm$}5.79 \\
& &DDPG(SimSR)+SAR &10.11\scriptsize{$\pm$2.28} &13.69\scriptsize{$\pm$6.81} &16.20\scriptsize{$\pm$2.80} &14.04\scriptsize{$\pm$5.44} &\textbf{12.63}\scriptsize{$\pm$3.89} &12.63\scriptsize{$\pm$3.89} \\
&                          
& GoFar
&18.38\scriptsize{$\pm$}3.24 
&16.11\scriptsize{$\pm$}2.39
&15.45\scriptsize{$\pm$}2.71
&13.66\scriptsize{$\pm$}3.97
&\textbf{8.19}\scriptsize{$\pm$}3.99
&10.57\scriptsize{$\pm$}3.93\\
&
&GoFar+SCAA 
&15.13\scriptsize{$\pm$}3.23 &16.03\scriptsize{$\pm$}2.44
&16.83\scriptsize{$\pm$}2.59
&16.37\scriptsize{$\pm$}2.87
&8.26\scriptsize{$\pm$}3.66
&\textbf{8.00}\scriptsize{$\pm$}3.76 \\
& &GoFar(SimSR) &14.19\scriptsize{$\pm$}4.49
&13.13\scriptsize{$\pm$}4.68 
&15.52\scriptsize{$\pm$}3.88 
&13.91\scriptsize{$\pm$}4.49 
&\textbf{12.44}\scriptsize{$\pm$}3.80 
&13.08\scriptsize{$\pm$}4.08
                            \\
& &GoFar(SimSR)+SAR
&14.19\scriptsize{$\pm$}4.49
&\textbf{12.44}\scriptsize{$\pm$}2.39
&16.08\scriptsize{$\pm$}2.36
&14.78\scriptsize{$\pm$}2.23
&14.19\scriptsize{$\pm$}4.49
&14.19\scriptsize{$\pm$}4.49
\\
                            \midrule
\multirow{8}{*}{FetchReach}  & \multirow{8}{*}{0.1}     & DDPG                    &29.92\scriptsize{$\pm$0.20}      &28.86\scriptsize{$\pm$0.46}      &24.21\scriptsize{$\pm$2.21}       &20.14\scriptsize{$\pm$5.04}        &2.02\scriptsize{$\pm$3.32}   &\textbf{1.86}\scriptsize{$\pm$3.31}      \\

& &DDPG+SCAA &29.88\scriptsize{$\pm$0.14} &29.08\scriptsize{$\pm$0.31} &28.85\scriptsize{$\pm$0.39} 
&29.36\scriptsize{$\pm$0.21}
&\textbf{27.68}\scriptsize{$\pm$1.30} &27.96\scriptsize{$\pm$0.91}       \\
& &DDPG(SimSR) &29.90\scriptsize{$\pm$0.18} 
&29.46\scriptsize{$\pm$}0.40
&28.68\scriptsize{$\pm$}0.97
&26.76\scriptsize{$\pm$}0.98
&27.72\scriptsize{$\pm$}2.35
&\textbf{27.16}\scriptsize{$\pm$}2.75 \\
& &DDPG(SimSR)+SAR &29.91\scriptsize{$\pm$0.20} &29.29\scriptsize{$\pm$0.43}
&28.47\scriptsize{$\pm$0.97}
&\textbf{27.53}\scriptsize{$\pm$1.70}
&29.91\scriptsize{$\pm$0.20}
&29.91\scriptsize{$\pm$0.20}\\
&                          & GoFar
& 27.84\scriptsize{$\pm$0.63}
&27.57\scriptsize{$\pm$0.92}
&27.22\scriptsize{$\pm$0.49}
&27.22\scriptsize{$\pm$0.49}
&15.71\scriptsize{$\pm$4.48}
&\textbf{11.24}\scriptsize{$\pm$6.76}
                            \\
&
&GoFar+SCAA 
&27.91\scriptsize{$\pm$0.36}
&27.72\scriptsize{$\pm$0.27}
&27.93\scriptsize{$\pm$0.35} 
&27.93\scriptsize{$\pm$0.35}
&15.70\scriptsize{$\pm$5.03} &\textbf{13.99}\scriptsize{$\pm$6.46}\\
& &GoFar(SimSR) &27.93\scriptsize{$\pm$}0.34 
&\textbf{27.91}\scriptsize{$\pm$}0.34 
&27.95\scriptsize{$\pm$}0.33 &27.96\scriptsize{$\pm$}0.32 
&27.96\scriptsize{$\pm$}0.29 
&27.98\scriptsize{$\pm$}0.33 \\
& &GoFar(SimSR)+SAR
&27.93\scriptsize{$\pm$}0.34
&\textbf{27.84}\scriptsize{$\pm$}0.32
&27.96\scriptsize{$\pm$}0.37
&27.98\scriptsize{$\pm$}0.36
&27.93\scriptsize{$\pm$}0.34
&27.93\scriptsize{$\pm$}0.34
\\
                            \midrule
\multirow{8}{*}{FetchSlide} & \multirow{8}{*}{0.1}     & DDPG                    &0.58\scriptsize{$\pm$1.30} &1.89\scriptsize{$\pm$0.89} &0.59\scriptsize{$\pm$1.09} &0.50\scriptsize{$\pm$1.12}  &\textbf{0.00}\scriptsize{$\pm$0.00}
&\textbf{0.00}\scriptsize{$\pm$0.00}
\\
& &DDPG+SCAA &0.74\scriptsize{$\pm$1.57} &1.54\scriptsize{$\pm$1.21} &1.08\scriptsize{$\pm$1.02} &0.94\scriptsize{$\pm$1.36} &\textbf{0.00}\scriptsize{$\pm$0.43} &0.18\scriptsize{$\pm$0.85} \\ 
& &DDPG(SimSR) &0.62\scriptsize{$\pm$}0.54 &0.83\scriptsize{$\pm$}0.77 &0.80\scriptsize{$\pm$}0.52 &0.43\scriptsize{$\pm$}0.90 &\textbf{0.37}\scriptsize{$\pm$}0.60 &0.84\scriptsize{$\pm$}0.71 \\
& &DDPG(SimSR)+SAR 
&0.62\scriptsize{$\pm$}0.54
&0.61\scriptsize{$\pm$}0.73
&\textbf{0.55}\scriptsize{$\pm$}0.62
&0.63\scriptsize{$\pm$}1.02
&0.62\scriptsize{$\pm$}0.54
&0.62\scriptsize{$\pm$}0.54
\\
&   & GoFar
&2.55\scriptsize{$\pm$1.37}
&1.23\scriptsize{$\pm$1.14}
&0.77\scriptsize{$\pm$0.92}
&1.13\scriptsize{$\pm$0.72}
&\textbf{0.00}\scriptsize{$\pm$0.00}
&\textbf{0.00}\scriptsize{$\pm$0.00}
                            \\ 
& &GoFar+SCAA 
&1.12\scriptsize{$\pm$0.95} &1.13\scriptsize{$\pm$0.91}
&0.90\scriptsize{$\pm$0.79}
&0.86\scriptsize{$\pm$0.84} &0.27\scriptsize{$\pm$0.61} &\textbf{0.00}\scriptsize{$\pm$0.00}
                            \\
& &GoFar(SimSR) &1.66\scriptsize{$\pm$}0.75 
&1.78\scriptsize{$\pm$}0.78 
&2.31\scriptsize{$\pm$}1.36 
&1.41\scriptsize{$\pm$}0.92 
&1.07\scriptsize{$\pm$}1.13 
&\textbf{0.59}\scriptsize{$\pm$}0.66
                            \\
& &GoFar(SimSR)+SAR  
&1.66\scriptsize{$\pm$}0.75
&1.95\scriptsize{$\pm$}1.30
&\textbf{1.47}\scriptsize{$\pm$}0.79
&2.25\scriptsize{$\pm$}1.29
&1.66\scriptsize{$\pm$}0.75
&1.66\scriptsize{$\pm$}0.75 
\\
\bottomrule
\end{tabular}